\documentclass[10pt,twocolumn,letterpaper]{article}

\usepackage{cvpr}
\usepackage{times}
\usepackage{epsfig}
\usepackage{graphicx}
\usepackage{amsmath}
\usepackage{amssymb}
\usepackage{multirow}

\usepackage[pagebackref=true,breaklinks=true,letterpaper=true,colorlinks,bookmarks=false]{hyperref}

\cvprfinalcopy 

\def\cvprPaperID{6696} 

\ifcvprfinal\fi
\begin{document}
	
	\title{Monocular 3D Object Detection Leveraging Accurate Proposals \\ and Shape Reconstruction}
	
	\author{Jason Ku\thanks{Equal contribution.} \quad Alex D. Pon$^\ast$ \quad Steven L. Waslander \\
	University of Toronto \\
	\tt\small\{kujason.ku, alex.pon\}@mail.utoronto.ca, stevenw@utias.utoronto.ca
	}
	
	\maketitle
	
	\begin{abstract}
		We present MonoPSR, a monocular 3D object detection method that leverages proposals and shape reconstruction. First, using the fundamental relations of a pinhole camera model, detections from a mature 2D object detector are used to generate a 3D proposal per object in a scene. The 3D location of these proposals prove to be quite accurate, which greatly reduces the difficulty of regressing the final 3D bounding box detection. Simultaneously, a point cloud is predicted in an object centered coordinate system to learn local scale and shape  information. However, the key challenge is how to exploit shape information to guide 3D localization. As such, we devise aggregate losses, including a novel projection alignment loss, to jointly optimize these tasks in the neural network to improve 3D localization accuracy. We validate our method on the KITTI benchmark where we set new state-of-the-art results among published monocular methods, including the harder pedestrian and cyclist classes, while maintaining efficient run-time.
	\end{abstract}
	
	\section{Introduction}
	
	A cornerstone of 3D scene understanding in computer vision is 3D object detection---the task where objects of interest within a scene are classified and identified by their 6 DoF pose and dimensions. Existing methods vary in the data they use, which include LiDAR~\cite{ku_avod, qi_fpointnet, yan2018second, pixor, zhou}, stereo images~\cite{chen_3dop}, and monocular images~\cite{chabot_deepmanta, lindernoren, kundu_3drcnn, mousavian_deep3dbox, xu_multifusion}. Monocular methods are attractive as they have the lowest cost and the simplest setup, relying on only a single camera. These methods are therefore attractive for applications where resources are limited, or for companies wanting to bring 3D object detection to mass markets such as autonomous navigation and virtual reality.
	
	Monocular 3D object detection methods are also the most disadvantaged; the problem formulation is under-constrained because depth information is lost when a 3D scene is projected onto an image plane. The difficulty of the problem is highlighted on the KITTI 3D Object Detection benchmark~\cite{geiger_kitti} in the car category where the best published monocular method~\cite{lindernoren} has an AP value $67\%$ lower than the best published method using LiDAR~\cite{yan2018second}. Results for the more challenging pedestrian and cyclist classes are rarely reported for monocular methods, likely due to even poorer performance.
	
	\begin{figure}[t!]
		\begin{center}
			\includegraphics[width=1.0\linewidth]{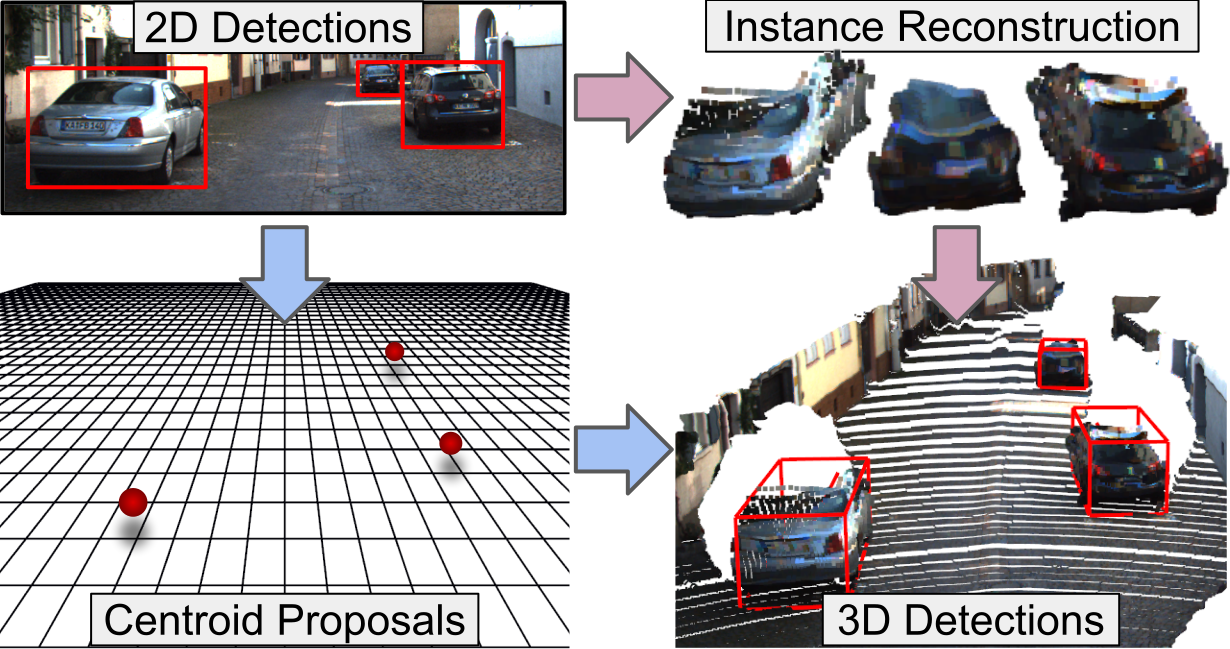}
		\end{center}
		\caption{\textbf{Pipeline for 3D Object Detection and Instance Point Cloud Estimation:} Our network takes an image with 2D bounding boxes and regresses instance-centric 3D proposals to produce 3D bounding boxes. Simultaneously, instance point clouds are estimated to recover local shape and scale, and to enforce 2D-3D consistency. The proposal regression and point cloud estimation are trained jointly in the network.}
		\label{fig:pipeline}
	\end{figure}
	
	To deal with the under-constrained monocular object detection problem, recent methods have typically used deep learning with well-informed priors. One such prior is that the predicted 3D bounding box should fit tightly within its corresponding 2D bounding box~\cite{mousavian_deep3dbox}. This assumption, however, leads to localization inaccuracies because it causes 2D bounding box, orientation, and dimension estimation errors to propagate to the final 3D box prediction. Other methods~\cite{chabot_deepmanta, kundu_3drcnn} match objects to CAD models to determine shape and pose. Unfortunately, these methods are restricted to the shape space covered by the selected CAD models, and do not easily extend to applications where models are unavailable. Lastly, all state-of-the-art methods under-utilize the information available during training. Although depth maps or LiDAR scans are available because they are required to create 3D labels, only \cite{xu_multifusion} incorporates this form of depth information during training, but neglect to exploit the strong priors that can be formed from 2D bounding boxes, such as the ones used by~\cite{lindernoren, mousavian_deep3dbox}.
	
	This paper introduces a proposal based monocular 3D object detection method that leverages the related task of shape reconstruction. We first greatly reduce the 3D search space by exploiting the robust performance of a 2D object detector by designing a non-restrictive 3D bounding box proposal per object detected in the scene. The location of the proposal is determined by considering the re-projection of the box center, and the relation between the height and depth of an object in the perspective transformation of a pinhole camera model. Compared to the 2D box constraint proposed by \cite{mousavian_deep3dbox}, our usage of a 2D bounding box does not lock in 2D box, orientation and dimensions inaccuracies. Instead, we use a two stage proposal regression design, in a similar manner to Faster R-CNN~\cite{ren_fasterrcnn}, which facilitates learning by regressing distributed anchor boxes, to obtain the final amodal, oriented 3D bounding box. We find this prior is flexible and suitable for classes that have varying dimensions and poses such as pedestrians and cyclists.
	
	We also incorporate an Instance Reconstruction module that predicts a point cloud for each instance in a canonical object (local) coordinate system. However, it is not obvious how to gain localization information from this estimated object point cloud. In our formulation, we connect the tasks of object detection and shape reconstruction by transforming the object point cloud into the camera coordinate frame using the instance centroid regression output. We finally jointly optimize the local scale and shape of each instance with its localization in the scene through multi-task learning and a novel projection alignment loss. This loss projects the object point cloud to image space and enforces 2D-3D consistency between the object point cloud and the image. The outputs for the proposal and shape reconstruction tasks are also designed to have smaller ranges, which has been shown to make learning tasks easier in various applications \cite{mousavian_deep3dbox, qi_fpointnet, ren_fasterrcnn}. An overview is provided in Fig.~\ref{fig:pipeline}.
	
	We validate our method on the KITTI 3D Object Detection benchmark \cite{geiger_kitti} on the car, pedestrian, and cyclist categories, and perform extensive ablation studies to evaluate our design choices. In summary, our key contributions are: a) an effective non-restrictive incorporation of a 2D bounding box prior to generate high quality 3D centroid proposals; b) an instance reconstruction module, which helps in recovering the shape and localization of objects; c) a novel loss formulation to jointly optimize point cloud estimates in both object and camera coordinate frames to enforce consistency between the 2D and 3D estimations. Furthermore, we are the first to propose a learning method that jointly optimizes point cloud reconstruction and observation consistency to achieve accurate 3D localization.
	
	These contributions lead to state-of-the-art results on the KITTI 3D Object Detection benchmark where we achieve a $68\%$ increase over the previous monocular state-of-the-art \cite{lindernoren}. In addition, we are the first to publish 3D pedestrian and cyclist results on the test benchmark and achieve highly promising results.
	
	\begin{figure*}[t]
		\begin{center}
			\includegraphics[width=1.0\linewidth]{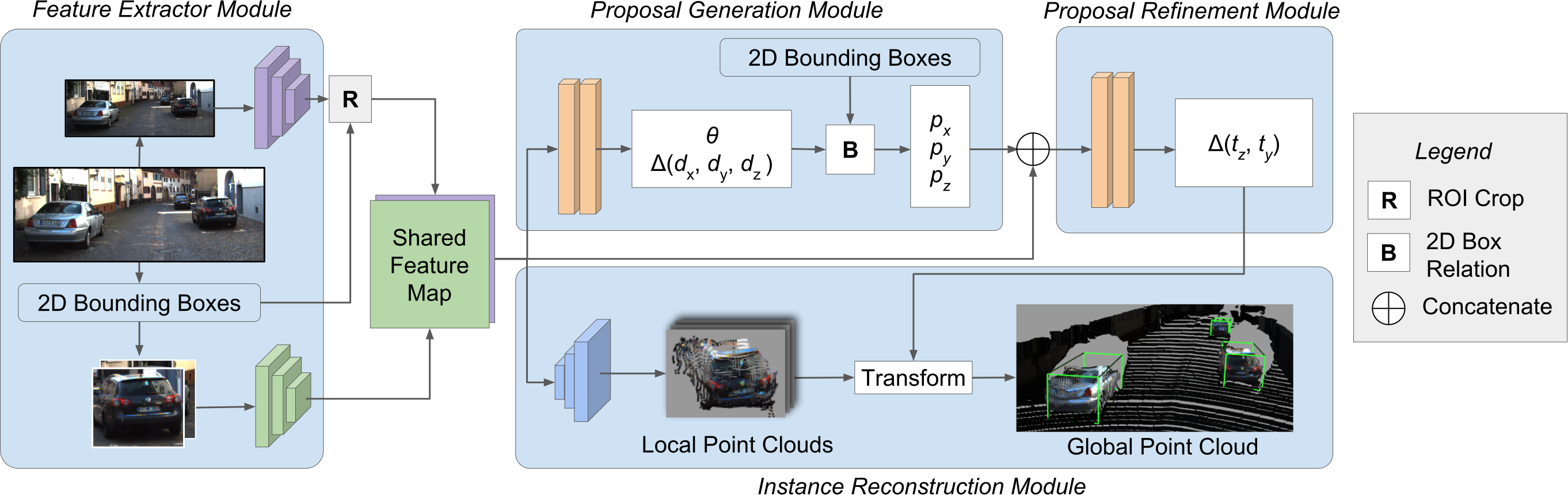}
		\end{center}
		\caption{\textbf{Network Architecture:} The network produces a feature map using an image crop of an object and global context features as inputs. From this feature map three tasks are performed a) the dimensions and orientation are predicted to estimate a proposal b) offsets for the proposals are regressed c) local point clouds are predicted and transformed into the global frame for auxiliary loss calculations.}
		\label{fig:architecture}
	\end{figure*}
	
	\section{Related Work}
	
	\paragraph{Proposal Based Methods}
	Many successful 2D object detectors employ the use of proposals~\cite{cheng2014bing, lee2015learning, ren_fasterrcnn, van2011segmentation, zitnick2014edge} to generate candidate object positions. This idea is extended into 3D scenes by methods such as Mono3D~\cite{chen_mono3d} and 3DOP~\cite{chen_3dop} which generate a large number of 3D proposals along an estimated ground plane, which are then projected to the image and scored by hand-crafted semantic, contextual, and shape features. Using 2D detections to reduce the search space in 3D has also shown promise because 2D detection is a mature field with robust performance; \cite{data_distillation_radosavovic} even suggests 2D detectors are accurate enough that detectors can be trained using data it inferences. $\text{F-PointNet}$~\cite{qi_fpointnet} lifts 2D detections to frustums and uses PointNets~\cite{qi2017pointnet, qi2017pointnet++} on these frustum points to regress 3D detections. In our monocular case, LiDAR point cloud information is not available. We instead use a pinhole model and leverage the relation of the 2D bounding box height and estimated object height to create centroid proposals that are regressed in the second stage of our network. 
	
	\paragraph{Geometric Priors}
	Geometry can be used as a prior to constrain the object detection problem. Deep3DBox~\cite{mousavian_deep3dbox} uses a neural network to predict the dimensions and pose of an object, then use a linear system of equations to enforce a constraint that the projected 3D bounding box must fit tightly in the 2D box. However, this hard constraint locks in errors from 2D bounding box, orientation, and dimension estimates when producing the 3D box. A3DODWTDA~\cite{lindernoren} instead estimate the image coordinates of the 3D bounding box corners, and solve a non-linear least squares fit for the best 3D box. While these methods work well for objects that maintain a constant shape like cars, solving for the projected 3D bounding box corners is harder for classes such as pedestrians; the 3D box dimensions and the corresponding projection of its corners vary greatly based on the skeletal pose of the person. In contrast, our formulation is less restrictive and does not lock in 2D bounding box, orientation, and dimension errors. While we still use geometry to generate a proposal, errors can be corrected in a later regression stage.
	
	\paragraph{Shape Reconstruction}
	Large-scale synthetic datasets such as ShapeNet~\cite{chang2015shapenet} have allowed deep neural networks to be trained on the task of shape reconstruction from single images~\cite{choy20163d, girdhar2016learning, kanazawa2016warpnet, kar2015category, soltani2017synthesizing, tulsiani2017multi, wu2016learning}. While shape reconstruction can be considered a separate problem, understanding shape can be beneficial for 3D object detection. Chabot et al.~\cite{chabot_deepmanta} use a network to output a 2D bounding box, vehicle part coordinates, and 3D box dimensions. They then match the dimensions to a CAD model and estimate pose using the matched model and the predicted vehicle parts. Kundu et al.~\cite{kundu_3drcnn} match CAD models using the shape space created by a set of CAD models, and train the network with a render-and-compare loss. While these methods have been shown to improve pose estimation, they require annotated datasets of 3D models for training, and can introduce error from CAD model mismatch. We instead devise a more flexible, class-agnostic solution that works with object point clouds directly and automatically generates relevant shape data from real-world LiDAR data to facilitate local shape learning.
	
	Moreover, methods involving shape completion \cite{dai2017shape, firman2016structured, rock2015completing} demonstrate the importance of enforcing consistency between the 3D estimations and 2D observations. Examples of differentiable 2D-3D consistency constraints for training deep networks are introduced in \cite{rezende2016unsupervised, tulsiani2017multi, wu2017marrnet, wu2016single, yan2016perspective}. The monocular 3D object detection methods \cite{lindernoren, mousavian_deep3dbox} loosely capture 2D-3D consistency when using geometric constraints with 2D boxes and the corners of 3D boxes. However, they ignore the shape of the object within the box. We, on the other hand, use the 3D point cloud of the object, and enforce 2D-3D consistency through a differentiable pixel-wise projection alignment loss.
	
	\paragraph{Depth Prediction}
	Recently, deep learning methods have shown significant improvement on the task of monocular depth prediction \cite{fu2018dorn, godard, kuznietsov2017semi}. MultiFusion \cite{xu_multifusion} uses depth prediction outputs from MonoDepth \cite{godard} and fuses this information with the corresponding RGB image to produce 3D bounding box estimates through a modified Faster R-CNN network. As opposed to predicting the depth of the entire scene, we use an instance-centric focus to make the task easier by avoiding regressing large depth ranges. We capture both shape and depth information in our formulation which predicts a point cloud in a local object frame then transforms it into the camera coordinate frame. In addition, we take a multi-task learning approach by sharing feature extractor weights used for box regression and point cloud estimation, and we combine our depth estimates with a well informed 2D box prior.
	
	\section{Monocular 3D Detection Framework}
	Given an $M \times N$ image $I$, the objective is to classify and localize objects of interest by fitting a 3D bounding box parameterized by its class $C$, centroid $T = (t_x, t_y, t_z)$, dimensions $D = (d_x, d_y, d_z)$, and orientation $O = (\theta, \phi, \psi)$. The core idea of our method is to reduce the search space by using a single high-quality proposal per object and to leverage shape reconstruction for accurate localization.
	
	We take advantage of the robust performance of existing 2D detectors to generate classified 2D bounding boxes. Using these boxes, image crops are passed through an encoder to output a feature map shared by the three downstream modules shown in Fig.~\ref{fig:architecture}: Proposal Generation, Proposal Refinement, and Instance Reconstruction. Full image features are included to provide additional scene context, and further improves results as shown in Sec.~\ref{sec:proposal_comparison}. 
	
	The objective of the Proposal Generation module is to generate high quality 3D proposals parameterized by their centroids $P = (p_x, p_y, p_z)$. The Proposal Refinement module regresses these proposals and outputs amodal, oriented 3D bounding boxes. The Instance Reconstruction module estimates a point cloud per instance in a local object coordinate frame, then transforms it into the camera coordinate frame using the centroid regression from the Proposal Refinement module. These modules are jointly optimized with a projection alignment loss.
	
	\subsection{Proposal Generation}\label{sec:proposal_generation}
	Features are extracted from the image using two encoders. The first encoder extracts features from the resized RGB crops of each instance detected by the 2D detector. The second encoder extracts features from the full image resized to half of its original size, and then RoI pooling \cite{girshick_fast} is used to crop a feature map for each instance. A shared feature map is generated through the depth-wise concatenation of both feature maps.
	
	Using the shared feature map, a proposal is generated per detected 2D bounding box as follows. For each 2D box, the orientation and dimensions are first estimated. The proposal depth is initialized from the perspective transformation relation of the object height in 3D and its projected height in image space. Lastly, the vertical and horizontal location are predicted, which are a function of the initialized depth and the horizontal and vertical viewing angles, $\alpha_h$ and $\alpha_v$, which are the rotations between the camera principle axis $[0, 0, 1]^T$ and the ray passing through the center of the 2D box.
	
	\paragraph{Orientation and Dimensions}
	In the KITTI benchmark implementation of this network, only the observation angle $\beta$ is estimated, which is the sum of the viewing angle $\alpha_h$ and object yaw $\theta$. As explained in \cite{mousavian_deep3dbox}, the estimation of an object's observation angle, instead of yaw, accounts for the change in appearance based on the viewing angle to the object. Both $\phi$ and $\psi$ are assumed to be zero in the KITTI benchmark, although the network can be extended to estimate them. The observation angle is estimated using discrete bins as in \cite{qi_fpointnet}. Dimensions offsets, $(\Delta d_x, \Delta d_y, \Delta d_z)$, are predicted from the mean class sizes. Using discrete bins and predicting offsets from mean sizes facilitates orientation and dimension learning by restricting values to be within a smaller range. These estimations are predicted early in the network to be made available for proposal initialization and refinement.
	
	\paragraph{Proposal Depth} The depth of an object is the most challenging parameter to estimate due to the large range in expected values, which can range from 5 m to 80 m on the KITTI dataset, and the fact that this information is lost during perspective projection. The classical pinhole camera model provides a relation between object height $h$, depth from the image plane $t_z$, focal length $f$, and projected height on the image plane $\hat{h}$ through similar triangles,
	
	\begin{equation}
	t_z = f \frac{h}{\hat{h}}.
	\label{eq:pin_hole}
	\end{equation}
	We can use this relation with object height estimates to predict the proposal depth $p_z$. It is important to note that the height, $h$, and the actual object height, ${d_y}$, are rarely equivalent due to perspective projection and camera viewpoint. However, for a camera with a viewpoint approximately parallel to the ground using this approximation provides a reasonable initial depth estimation. The depth of the proposal is initialized to the value calculated from Eq.~\ref{eq:pin_hole}, and we show in Sec.~\ref{sec:experiments} that this initialization provides a more accurate estimate of an object's depth compared to a direct estimation from the network.
	
	\paragraph{Proposal Vertical and Horizontal Location}
	The horizontal and vertical location $(p_x, p_y)$ of the proposal is determined by re-projecting the center coordinate of the 2D box $(u_c, v_c)$ into 3D space at depth $z_c = p_z$ using the camera calibration. The corresponding 3D point in camera coordinates is $(x_c, y_c, z_c)$ where $x_c$ and $y_c$ are the horizontal and vertical locations, respectively.
	
	\subsection{Proposal Refinement}
	The Proposal Refinement module further regresses the proposal location initialized in the previous stage of the network. Mousavian et al. \cite{mousavian_deep3dbox} choose to regress box dimensions $D$, rather than the translations, $T$, because there is smaller variance in box dimensions, which improves regression performance. We also make the learning task easier by regressing values within a small range, but since we generate an accurate proposal in the previous step, we formulate the refinement regression as an offset from the proposal, which we show provides better results than directly estimating depth in Sec.~\ref{sec:experiments}. The proposal depth error can be modeled as a function $s$ in,
	
	\begin{equation}
	\begin{aligned}
	t_z = f \frac{{{d_y}}}{b_{h}} + s(\theta, \phi, \psi, D, \alpha_v, \alpha_h)
	\end{aligned}
	\label{eq: network_pin_hole}
	\end{equation}
	where $s$ is dependent on the rotation of the object, its dimensions, and the viewing angles $\alpha_v, \alpha_h$. The first term of Eq.~\ref{eq: network_pin_hole} is the proposal depth $p_z$ which uses object height $d_y$ as $h$ and the 2D bounding box height $b_{h}$ as $\hat{h}$. We also confirm that learning the regression from the proposal, as opposed to a direct estimation, provides a more stable error in depth in Sec.~\ref{sec:experiments}.
	
	In the network, the shared feature maps from the feature extractor are flattened and concatenated with the proposed centroid, 2D box size, dimensions, and orientation. This feature vector is then passed through two sets of fully connected layers that output the translation regression targets $(\Delta t_y, \Delta t_z)$. Instead of regressing $\Delta t_x$, which would be overconstraining the problem, $t_x$ is recovered using the ray from the camera center to the estimated 3D centroid of the box.
	
	\subsection{Instance Reconstruction} \label{sec:instance_reconstruction}
\begin{figure}[t]
		\begin{center}
			\includegraphics[width=0.70\linewidth]{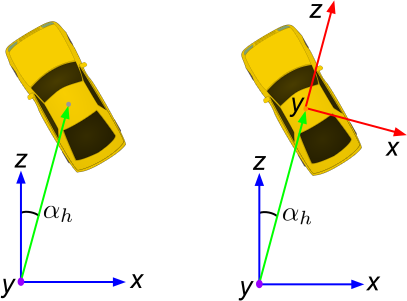}
		\end{center}
		\caption{\textbf{Object Coordinate System:} The instance point cloud is predicted within an object coordinate system created by translating the origin by the object centroid and rotating by the horizontal viewing angle $\alpha_h$.}
		\label{fig:pc_normalization}
	\end{figure}

	The Instance Reconstruction module takes advantage of the available LiDAR data during training to learn shape and scale information. We encode shape and scale information through a predicted local point cloud $p_O=\{p_i \in \mathbb{R}^3, i=1,\dots, K\}$, with $K$ as the number of points, and facilitate the learning task by using an object coordinate frame as shown in Fig.~\ref{fig:pc_normalization}. Using local coordinate frames have recently been shown to be effective in applications of 3D scene understanding \cite{qi_fpointnet, rematas_soccer}. In contrast to MultiFusion~\cite{xu_multifusion} where a global depth map is estimated up to 70 m, the point cloud in the object coordinate frame only contains values up to the size of the objects. Moreover, the rich point cloud representation is able to estimate instance shape and size while allowing for variations in 3D shape.
	
	We next note that the tasks of point cloud estimation and object localization are closely related. The estimated centroid and horizontal viewing angle are used to transform the predicted point cloud into the camera coordinate system. The predictions of both the local point cloud and the object position should be consistent with the object's appearance in image space. It is therefore intuitive to optimize these tasks jointly. This is achieved through a $Z$-channel loss and a projection alignment error loss, which penalizes the misalignment of the instance point cloud projected back into the image.
	
	\paragraph{Generating 3D Instance Training Data}
	The sparse LiDAR scan is first interpolated using \cite{ku2018defense} and converted into a 3 channel tensor $J$ using the provided camera calibration, with each pixel representing a point $(x, y, z)$. Points within ground truth boxes are considered as the set of points for an instance. These points are used to generate three additional sets of ground truth. First, the projection of these points in image space provides instance segmentation masks. Second, the third channel of $J$ corresponds to the instance depth map, which is masked and resized to the RGB crop size. Third, the local point cloud is generated. As shown in Fig.~\ref{fig:pc_normalization}, the rotation matrix $R_{\alpha_{h}}\in SO(3)$ is calculated using the horizontal viewing angle, and an object's point cloud in the camera coordinate frame is calculated in homogeneous coordinates as $p_{C}=T_{CO}*p_{O}$, where $T_{CO}$ is the transformation matrix
	\begin{equation}\label{eq:transform}
	T_{CO}=\begin{bmatrix}
	R_{\alpha_{h}} & T \\
	0 & 1 
	\end{bmatrix},
	\end{equation}
	and $T$ is the ground truth object centroid. The local point cloud is calculated by reversing this transformation. Note that we resize and encode the local and global instance point clouds as three channel tensors $L \times W \times 3$. Note, the segmented LiDAR input is not required during inference and is only used to generate training data.
	
	\paragraph{Point Cloud Estimation}
	The shared feature maps are fed into a small decoder network that produces the local $L \times W \times 3$ point clouds, with each grid element representing a point in the object coordinate system. Valid points are obtained by applying the automatically generated instance segmentation mask corresponding to the visible portion of the object.
	
	The predicted point clouds provide a rich form of 3D scene understanding not captured by methods that only output 3D bounding boxes, and avoids the inflexibility of a CAD dataset that certain methods \cite{chabot_deepmanta, kundu_3drcnn} require. This more flexible formulation should allow for generalization to a wide variety of objects. The network can also be trained to additionally output an $L \times W$ segmentation mask to allow the object point clouds to be used as signals for other tasks such as geometric appearance tracking.
	
	\subsection{Training Losses}\label{sec:losses}
	The network is trained using a multi-task loss defined by
	\begin{equation}
	\begin{aligned}
	L_{total} = L_{t} + L_{\theta} + L_{dim} + L_{c}.
	\end{aligned}
	\label{eq: losses}
	\end{equation}
	where $L_{t}$ is the regression loss for the difference between the true centroid and proposal centroid, $L_{\theta}$ is for the object's orientation, $L_{dim}$ is the offset regression loss for the bounding box dimensions, and $L_{c}$ represents the Instance Reconstruction losses described in the next section. The orientation loss, $L_{\theta}$, is formulated as in \cite{qi_fpointnet} with a classification loss for the discrete angle bins and a regression loss for the angle bin offsets. All regression losses use the smooth L1 loss and all classification losses use the softmax loss. Each loss is weighed such that validation losses converge to values with approximately the same magnitude.
	
	\subsubsection{Instance Reconstruction Losses}
\begin{figure}[t!]
		\begin{center}
			\includegraphics[width=1.0\linewidth]{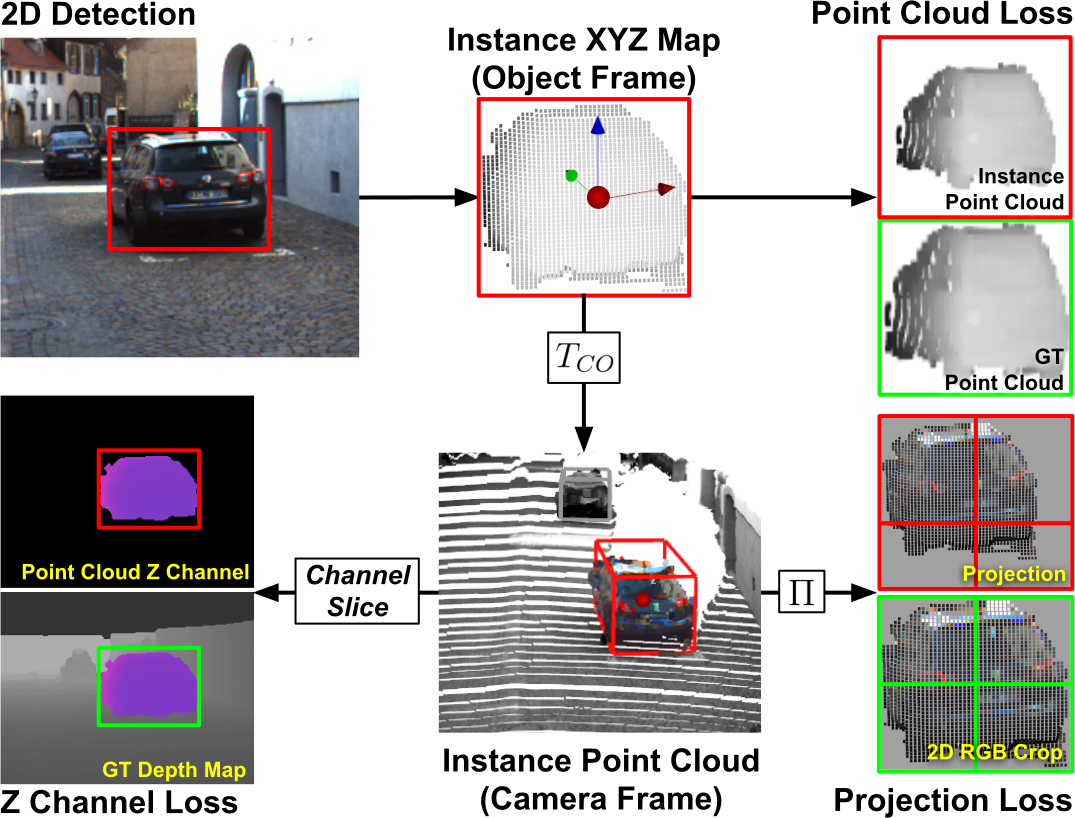}
		\end{center}
		\caption{\textbf{Instance Reconstruction Losses:} Losses for the corresponding predictions (\textbf{red}) and ground truth (\textbf{green}). All penalties use the smooth L1 loss at valid pixel locations using automatically generated segmentation masks. First, the point cloud loss penalizes the instance point cloud along each channel $(x,y,z)$. The point cloud is then placed at its estimated location in the camera coordinate frame using $T_{CO}$, the transformation between object and camera coordinate frames, and penalized in the last channel $z$. Finally, the point cloud is projected into image space with $\Pi$, the camera projection matrix. A projection alignment loss penalizes points projected into the wrong image pixel location.}
		\label{fig:hybrid_losses}
	\end{figure}

	Fig.~\ref{fig:hybrid_losses} shows the three losses used in the Instance Reconstruction module. The instance point cloud is represented as an $L \times W \times 3$ grid of points. Valid points, selected from the generated segmentation masks, are trained with a smooth L1 loss. The joint localization loss is composed of two parts, a $Z$-channel loss and a projection alignment loss.
	
	The local point cloud is transformed from its object coordinate frame to the camera coordinate frame using the transform $T_{CO}$ from Eq.~\ref{eq:transform}. The last channel of the transformed point cloud corresponds to the scene's depth map located at the instance, and is trained with a smooth L1 loss.
	
	The global instance point cloud can be projected into the image. From the generation process explained in Sec.~\ref{sec:instance_reconstruction} each projected point has a known image coordinate location within the 2D bounding box. The projected image coordinates $H$ are calculated using $H=\Pi p_{C}$ where $\Pi$ is the camera projection matrix. The projection alignment error is calculated with the expected coordinates $G$ as $E_{proj}=|G-H|$, and trained using a smooth L1 loss with regression targets of 0 at valid pixel locations. The projection error values are normalized by the width and height of the 2D bounding box. This loss penalizes the misalignment when the point cloud is projected to the image, which enforces the consistency between the 3D estimation and 2D appearance.
	
	\begin{table*}[t!]
		\small
		\centering
		\begin{tabular}{|l||c@{~/~}cc@{~/~}cc@{~/~}c||c@{~/~}cc@{~/~}cc@{~/~}c|}
			\hline
			\multicolumn{1}{|c||}{Method} & \multicolumn{6}{c||}{0.5 IoU} & \multicolumn{6}{c|}{0.7 IoU} \\
			\cline{2-13} &
			\multicolumn{2}{c}{Easy} & \multicolumn{2}{c}{Moderate} & \multicolumn{2}{c||}{Hard} &
			\multicolumn{2}{c}{Easy} & \multicolumn{2}{c}{Moderate} & \multicolumn{2}{c|}{Hard} \\ \hline
			Mono3D~\cite{chen_mono3d}   & 30.50 &   -   & 22.39 &   -   & 19.16 &   -   &  5.22 &   -   &  5.19 &   -   &  4.13 &   -   \\
			Deep3DBox~\cite{mousavian_deep3dbox}  &   -   & 30.02 &   -   & 23.77 &   -   & 18.83 &   -   &  9.99 &   -   &  7.71 &   -   &  5.30 \\ 
			A3DODWTDA~\cite{lindernoren}        & 45.46 &   -   & 33.83 &   -   & 31.78 &   -   & 15.64 &   -   & 12.90 &   -   & 12.30 &   -   \\
			MultiFusion~\cite{xu_multifusion} & 55.02 & 54.18 & 36.73 & 38.06 & 31.27 & 31.46 & \textbf{22.03} & 19.20 & 13.63 & 12.17 & 11.60 & 10.89 \\
			\hline
			Ours        & \textbf{56.97} & \textbf{55.45} & \textbf{43.39} & \textbf{43.31} & \textbf{36.00} & \textbf{35.47} &
			20.63  & \textbf{21.52} & \textbf{18.67} & \textbf{18.90} & \textbf{14.45} & \textbf{14.94} \\
			\hline
		\end{tabular}
		\caption{\textbf{3D Localization:} \emph{$AP_{BEV}$} on KITTI \emph{val1}/\emph{val2} sets.}
		\label{tab:kitti_val_bev}
	\end{table*}
	
	\begin{table*}[t!]
		\small
		\centering
		\begin{tabular}{|l||c@{~/~}cc@{~/~}cc@{~/~}c||c@{~/~}cc@{~/~}cc@{~/~}c|}
			\hline
			\multicolumn{1}{|c||}{Method} & \multicolumn{6}{c||}{0.5 IoU} & \multicolumn{6}{c|}{0.7 IoU} \\
			\cline{2-13} &
			\multicolumn{2}{c}{Easy} & \multicolumn{2}{c}{Moderate} & \multicolumn{2}{c||}{Hard} &
			\multicolumn{2}{c}{Easy} & \multicolumn{2}{c}{Moderate} & \multicolumn{2}{c|}{Hard} \\ \hline
			Mono3D~\cite{chen_mono3d}   & 25.19 &   -   & 18.20 &   -   & 15.52 &   -   &  2.53 &   -   &  2.31 &   -   &  2.31 &   -   \\
			Deep3DBox~\cite{mousavian_deep3dbox} &   -   & 27.04 &   -   & 20.55 &   -   & 15.88 &    -  &  5.85 &   -   &  4.10 &   -   &  3.84 \\ 
			A3DODWTDA~\cite{lindernoren}        & 40.31 &   -   & 30.77 &   -   & 26.55 &   -   & 10.13 &   -   &  8.32 &   -   &  8.20 &   -   \\
			MultiFusion~\cite{xu_multifusion}  & 47.88 & 44.57 & 29.48 & 30.03 & 26.44 & 23.95 & 10.53 &  7.85 &  5.69 &  5.39 &  5.39 &  4.73 \\ \hline
			Ours      & \textbf{49.65} &   \textbf{48.89}   & \textbf{41.71} &   \textbf{40.93}   & \textbf{29.95} & \textbf{33.43}   &  
			\textbf{12.75} &   \textbf{13.94}   & \textbf{11.48} &   \textbf{12.24}   & \textbf{8.59} &   \textbf{10.77} \\
			\hline
		\end{tabular}
		\caption{\textbf{3D Detection:} \emph{$AP_{3D}$} on KITTI \emph{val1}/\emph{val2} sets.}
		\label{tab:kitti_val_3d}
	\end{table*}
	
	\begin{table*}[t!]
		\small
		\centering
		\begin{tabular}{|l||c|ccc||ccc|}
			\hline
			\multicolumn{1}{|c||}{Method} & & \multicolumn{3}{c||}{BEV AP} & \multicolumn{3}{c|}{3D AP} \\
			\cline{3-8}               & Runtime &  Easy & Moderate &  Hard &  Easy & Moderate &  Hard \\ \hline
			MultiFusion~\cite{xu_multifusion} & 0.12 & 13.73 &     9.62 &  8.22 &  7.08 &     5.18 &  4.68 \\
			A3DODWTDA~\cite{lindernoren} & 0.80 & 10.21 &    10.61 &  8.64 &  6.76 &     6.45 &  4.87 \\ \hline
			Ours                         & 0.20 & \textbf{20.25} & \textbf{17.66} & \textbf{15.78} &
			\textbf{12.57} & \textbf{10.85} & \textbf{ 9.06} \\ \hline
		\end{tabular}
		\caption{\textbf{3D Car Localization and Detection:} \emph{$AP_{BEV}$} and \emph{$AP_{3D}$} AP on KITTI \emph{test} set.}
		\label{tab:kitti_test}
	\end{table*}
	
	\begin{table*}[t!]
		\small
		\centering
		\begin{tabular}{|c||ccc||ccc|}
			\hline
			Metric & \multicolumn{3}{c||}{Pedestrians} & \multicolumn{3}{c|}{Cyclists} \\
			\cline{2-7} & Easy & Moderate & Hard & Easy & Moderate & Hard \\
			\hline
			BEV AP    & 14.27 & 11.22 & 10.54 & 14.75 & 12.17 & 11.35 \\ \hline
			3D AP     & 12.65 & 10.66 & 10.08 & 13.43 & 11.01 &  9.93 \\ \hline
		\end{tabular}
		\caption{\textbf{3D Pedestrian and Cyclist Detection}: \emph{$AP_{BEV}$} and \emph{$AP_{3D}$} for the pedestrian and cyclist classes on the KITTI \emph{test} split. No other published method currently has results on the test server.}
		\label{tab:kitti_test_ped_cyc}
	\end{table*}
	
	\section{Implementation}
	We use MS-CNN \cite{cai} as our 2D detector for fair comparison with \cite{mousavian_deep3dbox}. To facilitate faster convergence, the full scene and instance convolution encoders are initialized with ResNet-101 weights before $\textit{conv}5\_1$ pre-trained on the task of 2D object detection on KITTI. The full image is resized to $160 \times 608$, and each instance is cropped and resized into a $48 \times 48 \times 3$ image from the RGB image. The feature extractor output stride is set to 4, resulting in a final layer feature map with resolution $12 \times 12$. A flattened version of this map is passed into two fully connected layers for box regression and orientation estimation. The local point cloud is generated from the small decoder network consisting of repeated upsampling and convolutional layers, resulting in a feature map of the original $48 \times 48$ resolution, after which a $48 \times 48 \times 3$ point cloud $p_l$ is outputted through a $3 \times 3$ convolutional layer. The network is trained using an Adam optimizer for 100K iterations with an initial learning rate of 0.0008 and decay factor of 0.8 every 10K steps.
	
	\begin{figure*}[t]
		\begin{center}
			\includegraphics[width=1.0\linewidth]{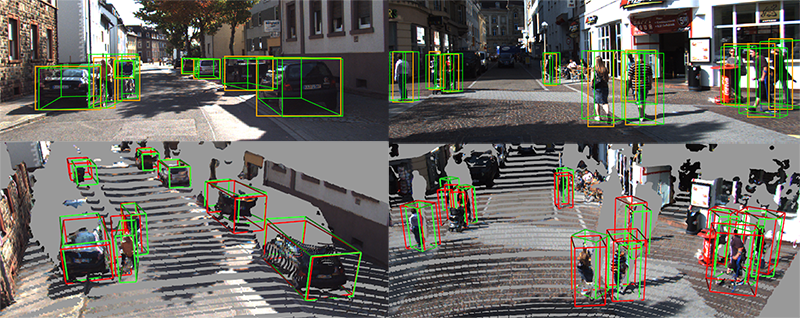}
		\end{center}
		\caption{Qualitative detection results on several scenes in the KITTI dataset. 2D detections (top) are shown in \textbf{orange}. 3D detections in \textbf{green} are shown projected into the image (top) and in the 3D scene (bottom). Ground truth 3D boxes (bottom) are shown in \textbf{red}. Points within the detection boxes are the estimated point clouds from the network, while the background points are taken from the colorized interpolated LiDAR scan. Note that for pedestrians in particular, the projected 3D boxes do not fit tightly within their 2D box, so constraining the 3D box with the 2D box is not ideal.}
		\label{fig:qual_results}
	\end{figure*}
	
	\section{Experiments}\label{sec:experiments}
	We present results on the challenging KITTI 3D Object Detection benchmark where we compare with the current state-of-the-art monocular methods, validate design choices through ablation studies, and present qualitative results. Two validation splits are used to compare against previous methods. The first split \textit{val1} follows \cite{chen_mv3d} and the second split \textit{val2} follows \cite{xiang_3dvp}. The ablation studies were performed on \textit{val1}. As with~\cite{lindernoren, qi_fpointnet}, a separate training split is used for better generalization on the test results. Training is done with a batch of up to 32 ground truth 2D boxes from the same image, while all inference and evaluation is done using the 2D detection boxes from MS-CNN \cite{cai}. The only data augmentation used during training is 2D box jittering to simulate 2D detections. Specifically, Gaussian noise was added to the dimensions and center of the 2D box, scaled by the size of the box, while keeping a minimum 0.7 IoU overlap with the original ground truth box. For evaluation, the KITTI easy, moderate, and hard difficulty classifications are used.
	
	\subsection{AP Comparison with State-of-the-Art Methods}
	We compare our approach with previous monocular state-of-the-art methods on the tasks of 3D localization and 3D detection, using the Bird's Eye View (BEV) and 3D AP metrics, on the KITTI validation splits in Tab.~\ref{tab:kitti_val_bev} and Tab.~\ref{tab:kitti_val_3d}, respectively. The results of~\cite{chen_mono3d, mousavian_deep3dbox} are taken from~\cite{xu_multifusion}. We also submit our detections to the KITTI test server for evaluation, with the results shown in Tab.~\ref{tab:kitti_test}. The results demonstrate that our method outperforms the previous state-of-the-art by a significant margin while maintaining efficient runtime. The total inference time for the network is 120ms on a Titan X GPU, which is in addition to the 2D detector which takes 80ms. We are also the first monocular 3D object detection method to publish pedestrian and cyclist results on the KITTI 3D Object Detection benchmark. Highly promising results on the test and validation sets are shown in Tab.~\ref{tab:kitti_test_ped_cyc} and Tab.~\ref{tab:kitti_val_ped_cyc}.
	
	\subsection{Effect of Proposals and Full Image Features} \label{sec:proposal_comparison}
	Tab.~\ref{tab:proposal_comparison} investigates the effect of the proposal formulation described in Sec.~\ref{sec:proposal_generation} and the importance of full image features for learning scene context. The focus of the proposal is to improve depth estimation, so the metrics used are average depth error and standard deviation.
	
	To show the viability of using a single proposal per detection, we train a model to directly estimate the object depth with full image features appended, and two models that predict regressions from proposals, one that regresses using only instance features and the other with full image features appended. The model trained for direct depth estimation performs worse than the proposal regression networks. The first two rows of Tab.~\ref{tab:proposal_comparison} show that proposal regression leads to more accurate centroid depth than the direct estimation method. The use of our proposals is further justified as they lead to a more stable error with lower standard deviation. The last row shows that the use of full image features provides useful cues for centroid depth estimation.

	\begin{table}[t!]
		\small
		\centering
		\begin{tabular}{|c|c|c||ccc|}
			\hline
			Class & Metric & IoU & Easy & Moderate & Hard\\
			\hline
			\multirow{4}{*}{Ped.} & \multirow{2}{*}{BEV AP} & 0.25 & 32.54 & 28.92 & 24.32 \\
			& & 0.5   & 11.68 & 10.05 & 8.14 \\ \cline{2-6}
			& \multirow{2}{*}{3D}     & 0.25 & 31.89 & 28.23 & 23.36 \\ 
			& & 0.5   & 10.64 & 8.18 & 7.18 \\ \hline
			\multirow{4}{*}{Cyc.} & \multirow{2}{*}{BEV AP} & 0.25 & 24.79 & 17.53 & 16.96 \\
			& & 0.5   & 11.18 & 10.18 & 10.03 \\ \cline{2-6}
			& \multirow{2}{*}{3D AP}  & 0.25 & 23.77 & 17.24 & 16.45 \\
			& & 0.5   & 10.88 & 9.93 & 9.93 \\ \hline 
			
		\end{tabular}
		\caption{\textbf{3D Pedestrian and Cyclist Detection}: \emph{$AP_{BEV}$} and \emph{$AP_{3D}$} for the pedestrian and cyclist classes on KITTI \emph{val1} set.}
		\label{tab:kitti_val_ped_cyc}
	\end{table}
	
	\subsection{Instance Reconstruction Analysis} \label{sec:instance_reconstruction_analysis}
	We analyze the effect of the Instance Reconstruction module in Tab.~\ref{tab:local_global}. The baseline network is taken as the proposal regression network using the full image from Sec.~\ref{sec:proposal_comparison}. We train three additional networks, and show the effect of each additional loss. The results show that the estimation of a local point cloud and its depth map are useful tasks for 3D object detection. The final row shows that the joint optimization of the losses through the projection alignment further helps the learning procedure, increasing $AP_{3D}$ performance at 0.5 IoU by 7.2\% over the baseline.

	\begin{table}[t]
		\small
		\centering
		\begin{tabular}{|l|c|c|}
			\hline
			\multicolumn{1}{|c|}{Version} & Input & Depth Error \\ \hline
			Proposal Only       & C+D   & 1.43 / 2.08  \\
			Proposal Regression & C+D   & 1.31 / 1.93 \\
			Direct Estimation   & C+F+D & 1.48 / 2.28 \\
			Proposal Regression & C+F+D & \textbf{1.22} / \textbf{1.84} \\
			\hline
		\end{tabular}
		\caption{\textbf{Proposal Representation and Full Image Features}. Error in meters for centroid depth estimation (average absolute error / standard deviation) for the hard car category on the KITTI \emph{val1} set. C = features from the RGB crop, D = the estimated 2D and 3D box dimensions, F = features from the full image. Networks here do not include the Instance Reconstruction module.}
		\label{tab:proposal_comparison}
	\end{table}
	
	\section{Qualitative Results}
	Fig.~\ref{fig:qual_results} shows qualitative detection results on scenes from the KITTI dataset. It can be noted that the projection of the 3D boxes of cars and cyclists often match the 2D detection boxes closely. However, for pedestrians, the projection of the 3D boxes varies greatly from the 2D box, and constraining the 3D box to fit the 2D box would result in poor localization. In our formulation, the less restrictive proposal regression method allows for accurate localization of different objects in the scene, including pedestrians. In addition, the estimated instance point clouds shown in the 3D view appear consistent with their appearance in the image.
	
	\begin{table}[t]
		\small
		\centering
		\begin{tabular}{|c|c|c||c|}
			\hline
			Local      &      Depth & Projection & $AP_{3D}$ \\
			\hline
			-          &          - &          - &     34.47 \\
			\checkmark &          - &          - &     39.33 \\
			\checkmark & \checkmark &          - &     41.06 \\
			\checkmark & \checkmark & \checkmark & \textbf{41.71} \\
			\hline
		\end{tabular}
		\caption{\textbf{Instance Reconstruction Analysis}: The effect of the local point cloud estimation and the depth and projection losses. Results are evaluated for $AP_{3D}$ at 0.5 IoU on the moderate car category.}
		\label{tab:local_global}
	\end{table}
	
	\section{Conclusion}
	To conclude, this work presents a monocular 3D object detection method that uses accurate proposals to reduce the search space and leverages shape reconstruction through a point cloud. Object centroid estimation is formulated as an offset regression from proposals generated by a well informed 2D bounding box prior, and object scale and shape is encoded through a predicted point cloud in a canonical object coordinate frame. Accurate localization is achieved through joint optimization of these tasks through a depth map and projection alignment loss. These innovations lead to state-of-the-art results on the KITTI 3D Object Detection benchmark while maintaining efficient runtime.
	
	{\small
		\bibliographystyle{ieee_fullname}
		\bibliography{root}
	}
	
	\clearpage

    \appendix
    \onecolumn
    
    \section*{Supplementary Material: Additional Qualitative Results}
\begin{figure*}[h!tb]
	\begin{center}
		\includegraphics[width=0.96\linewidth]{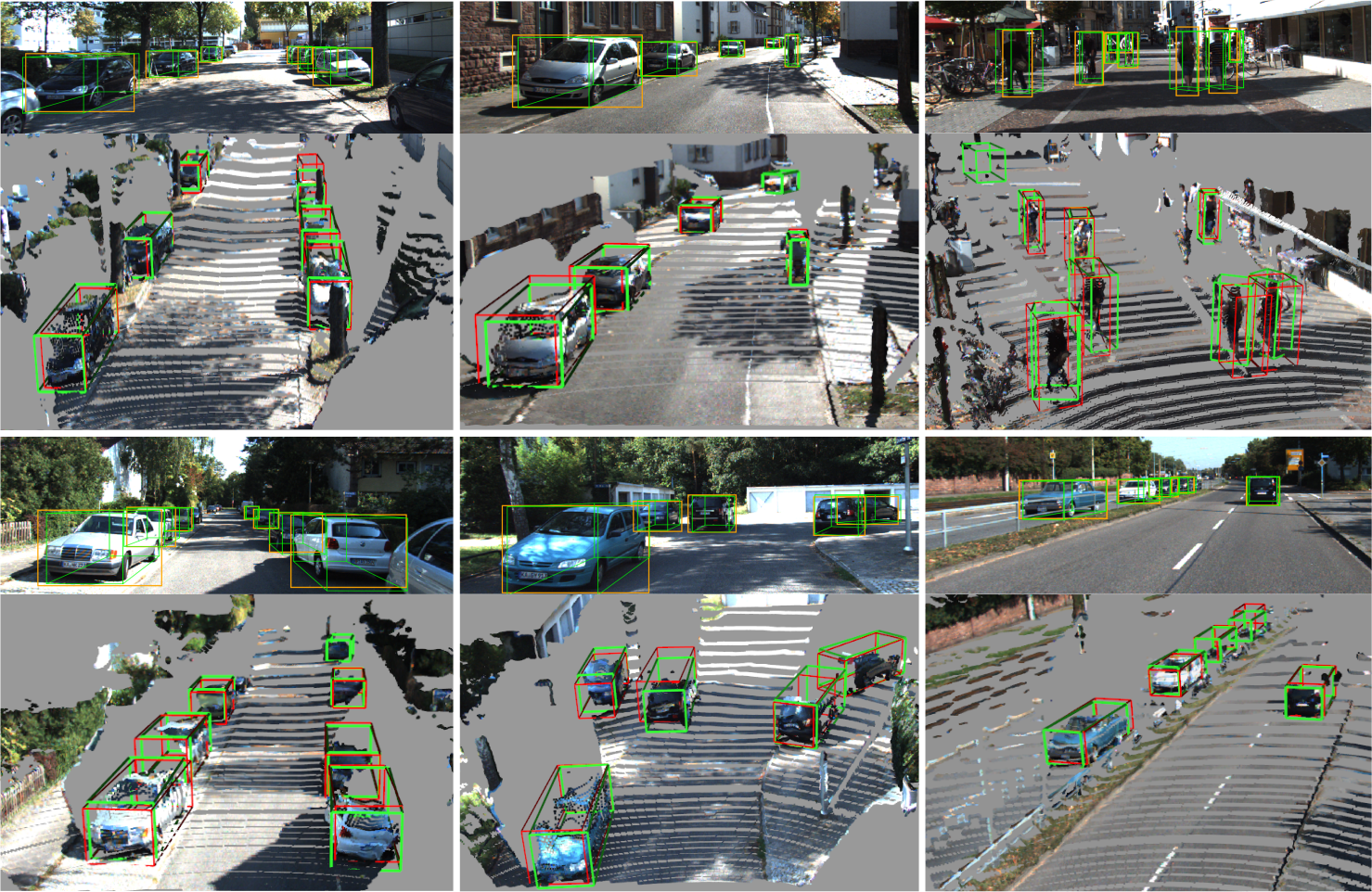}
	\end{center}
	\caption{\textbf{Additional Qualitative Results:} 2D detections (top) are shown in \textbf{orange}. 3D detections in \textbf{green} are shown projected into the image (top) and in the 3D scene (bottom). Ground truth 3D boxes (bottom) are shown in \textbf{red}. Points within the detection boxes are the estimated point clouds from the network, while the background points are taken from the colorized interpolated LiDAR scan.}
	\label{fig:supp_diagram}
	\end{figure*}

\begin{figure*}[h]
	\begin{center}
		\includegraphics[width=1.0\linewidth]{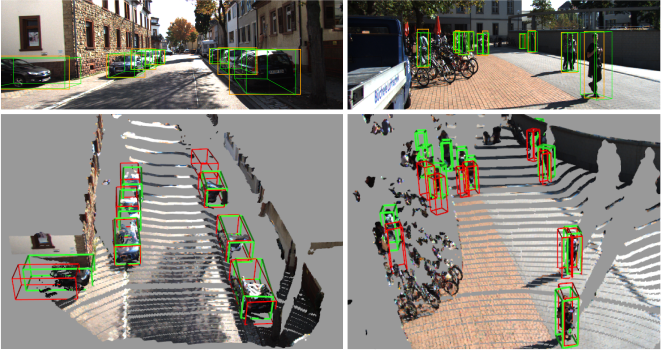}
	\end{center}
	\caption{\textbf{Common Failure Cases:} Truncation and occlusion are two common causes of localization errors. Truncation is evident for the far left car in the first image, and occlusion is common in large groups of pedestrians walking together as shown in the second image. 2D detections (top) are shown in \textbf{orange}. 3D detections in \textbf{green} are shown projected into the image (top) and in the 3D scene (bottom). Ground truth 3D boxes (bottom) are shown in \textbf{red}. Points within the detection boxes are the estimated point clouds from the network, while the background points are taken from the colorized interpolated LiDAR scan.}
	\label{fig:supp_diagram_2}
	\end{figure*}

	In Fig. \ref{fig:supp_diagram}, we show additional detection results on several scenes in the KITTI \cite{geiger_kitti} validation split, \textit{val1} \cite{chen_mv3d}. Fig. \ref{fig:supp_diagram_2} shows common failure cases for the network which are noted when there is heavy occlusion or truncation of the object. Additional results on several KITTI driving sequences can be found at \href{https://youtu.be/_iJpEpXB7j4}{https://youtu.be/\_iJpEpXB7j4}
	
	\addtolength{\textheight}{-9.7cm}
	
\end{document}